# DynamicPPL: Stan-like Speed for Dynamic Probabilistic Models

MOHAMED TAREK, University of New South Wales at Canberra, Australia
KAI XU, University of Edinburgh, UK
MARTIN TRAPP, Graz University of Technology, Austria
HONG GE, University of Cambridge, UK
ZOUBIN GHAHRAMANI, Uber AI Labs, USA and University of Cambridge, UK

We present the preliminary high-level design and features of `DynamicPPL.jl`[1], a modular library providing a lightning-fast infrastructure for probabilistic programming. Besides a computational performance that is often close to or better than `Stan`, `DynamicPPL` provides an intuitive DSL that allows the rapid development of complex dynamic probabilistic programs. Being entirely written in Julia, a high-level dynamic programming language for numerical computing, `DynamicPPL` inherits a rich set of features available through the Julia ecosystem. Since `DynamicPPL` is a modular, stand-alone library, any probabilistic programming system written in Julia, such as `Turing.jl`, can use `DynamicPPL` to specify models and trace their model parameters. The main features of `DynamicPPL` are: 1) a meta-programming based DSL for specifying dynamic models using an intuitive tilde-based notation; 2) a tracing data-structure for tracking RVs in dynamic probabilistic models; 3) a rich contextual dispatch system allowing tailored behaviour during model execution; and 4) a user-friendly syntax for probabilistic queries. Finally, we show in a variety of experiments that `DynamicPPL`, in combination with `Turing.jl`, achieves computational performance that is often close to or better than `Stan`.

## 1 INTRODUCTION

Probabilistic programming unifies traditional programming and probabilistic modelling in order to simplify the specification of probabilistic models. Developing probabilistic programming systems has a long history and is still a very active field of research [Bingham et al. 2019; Carpenter et al. 2017; Ge et al. 2018; Goodman et al. 2008; Kozen 1981; Lew et al. 2019; Lunn et al. 2000; Mansinghka et al. 2014, 2018; Milch et al. 2005; Minka and Winn 2008; Murray and Schön 2018; Pfeffer 2001, 2009; Wood et al. 2014]. One particularly relevant form of probabilistic programming is the so-called Bayesian probabilistic language (BPL), which extends probabilistic programming with automated Bayesian inference. Modern BPLs aim at 1) providing modern inference algorithms such as Hamiltonian Monte Carlo (HMC) [Neal et al. 2011], particle Markov Chain Monte Carlo (MCMC) [Andrieu et al. 2010], variational inference (VI) [Blei et al. 2017], message passing and customizable inference; 2) hardware acceleration exploiting parallelism, e.g. GPUs; and 3) expanding modelling families, e.g. neural networks and stochastic processes. This paper focuses on the computational efficiency aspect of dynamic BPLs, that is BPLs that support dynamic parameter types and dynamic model dimensionality. In particular, we present a system that automatically performs type inference for traces of dynamic probabilistic programs while utilising the type information to speed up future executions. This is achieved by exploiting the multiple dispatch and dynamic dispatch capabilities of the dynamically typed Julia programming language [Bezanson et al. 2017] to generate efficient machine code for Bayesian inference and model execution.

---

[1]https://github.com/TuringLang/DynamicPPL.jl

Authors' addresses: Mohamed Tarek, University of New South Wales at Canberra, Australia, m.mohamed@student.adfa.edu.au; Kai Xu, University of Edinburgh, UK, kai.xu@ed.ac.uk; Martin Trapp, Graz University of Technology, SPSC Lab, Austria, martin.trapp@tugraz.at; Hong Ge, University of Cambridge, CBL Lab, UK, hg344@cam.ac.uk; Zoubin Ghahramani, Uber AI Labs, USA, University of Cambridge, UK, zoubin@eng.cam.ac.uk.

Our main contributions are:
- A type inference method for dynamic traces which enables the generation of efficient machine code for Bayesian inference tasks.
- A dynamic tracing approach that facilitates Stan-like efficiency, while ensuring compatibility with Julia's automatic differentiation (AD) landscape.
- Contextual dispatch of model execution, which allows tailored behaviour during the model execution, e.g. sampling from custom proposal distributions.

## 1.1 Related Work

There is a large number of open-source probabilistic programming systems with different inference algorithm choices, levels of maturity, implementation languages and domain-specific language (DSL) designs. We do not attempt a complete review in this short paper. Instead, we focus on similar systems developed in the Julia language. However, the same set of comparison principles could be applied to evaluate other systems.

In addition to `DynamicPPL.jl`[2], the Julia programming language is seeing a surge of new BPLs being actively developed such as: Gen[3][Mansinghka et al. 2018], `Soss.jl`[4] and `ProbabilityModels.jl`[5] all of which are rapidly developing projects. Because of the fast development speed of these BPLs, we refrain from a detailed comparison of the features, implementation details and performance. However to the best of our knowledge, the philosophy of the two major BPLs, Gen and Turing, are actually more similar than different. Both Turing and Gen use a trace-based inference approach, in which models are annotated Julia functions and random variables (RVs) are given names or addresses at run-time. While Turing and Gen vary in their API and in the set of features they offer, to the best of our knowledge each feature available in Gen can be implemented in Turing and vice versa. Therefore, these differences in features will likely decrease as both projects keep maturing. Soss on the other hand takes a very different approach to BPL compared to both Turing and Gen. In Soss, the model is stored using a Julia abstract syntax tree (AST) [Bezanson et al. 2017] enabling model transformations and simplifications at the symbolic level before compilation. Fortunately, it is to be expected that the existing Julia BPLs will become interoperable at some point, so that users can mix and match between the features offered by each of these libraries.

## 2 A HIGH-PERFORMANCE, DYNAMIC LANGUAGE FOR PROBABILISTIC MODELS

We will briefly discuss the `DynamicPPL`-specific design in this section. We would like to point out that many of these techniques (see also e.g. [Lew et al. 2019; Ścibior and Thomas 2019]) could be implemented for other systems in the future.

### 2.1 Specifying Probabilistic Models

The key entry-point to specify any probabilistic model using `DynamicPPL` is the `@model` macro. Using the `@model` macro, the user can define a probabilistic model generator using an intuitive modelling syntax. The following examples illustrate the use of the `@model` macro to define generalized linear models, e.g. linear regression and logistic regression, using `DynamicPPL`.

---

[2]`DynamicPPL.jl` is a sub-module of `Turing.jl`, but can be used as an independent DSL implementation for other probabilistic programming systems. Similarly, other suitable DSLs, e.g. Gen or `Soss.jl`, can in principle be used as a DSL for `Turing.jl`, although not currently implemented.
[3]https://github.com/probcomp/Gen
[4]https://github.com/cscherrer/Soss.jl
[5]https://github.com/chriselrod/ProbabilityModels.jl

DynamicPPL: Stan-like Speed for Dynamic Probabilistic Models    3```
@model linreg(X, y) = begin          @model logreg(X, y) = begin
    d = size(X, 2)                       d = size(X, 2)
    w ~ MvNormal(zeros(d), 1)            w ~ MvNormal(zeros(d), 1)
    s ~ Gamma(1, 1)                      v = logistic.(X' * w)
    y .~ Normal.(X * w, s)               y .~ Bernoulli.(v)
end                                  end
```

Note that the dot notation, e.g. ".~", is Julia syntax for broadcasting. For example in the linreg model, each observation y[i] is independently normally distributed with mean (X * w)[i] and standard deviation s – c.f. last line. Running the above model definitions, will define linreg and logreg as instances of the model constructor type ModelGen. The user can then specify the values of the inputs (X, y) and construct an instance of Model by calling the respective model constructor, e.g. linreg(X, y), with a matrix of covariates X and a vector of observation y.

In DynamicPPL, RVs in the model are first identified. DynamicPPL then automatically determines the model parameters and data, i.e., observed RVs, during model construction. The determination of parameters is performed based on the types of the input arguments. More precisely, RVs which are not specified as input arguments to the model constructor and those given a value of missing will be treated as model parameters, which informs DynamicPPL to perform inference of the respective RVs. For each model parameter, an instance of VarName is constructed at run-time. Each VarName holds information about the user-specified variable symbol, e.g. "w", and additional indexing information in case of arrays.

### 2.2 Building Typed Traces for Dynamic Probabilistic Programs

In DynamicPPL, execution traces of all model parameters are stored in instances of so-called VarInfo types. Specifically, each instance of VarInfo associates the VarName of a RV to its current state, distribution as well as other metadata. VarInfo has two subtypes: UntypedVarInfo and TypedVarInfo. UntypedVarInfo uses a **Vector**{**Real**} to store the states of the model parameters, allowing for differently typed RVs to be stored in a single vector representation. Similarly, a **Vector**{Distribution} is used to store all the distributions of the respective RVs. Note that both **Real** and Distribution are abstract types in Julia thus allowing UntypedVarInfo to handle differently typed values but hindering the generation of efficient machine code by the Julia compiler. On the other hand, TypedVarInfo leverages a strictly typed vector representation of concrete Julia types, for both states and distributions, allowing the Julia compiler to generate highly efficient machine code for the sampling process and model execution. Having these two data structures at hand allows the use of UntypedVarInfo during the initial sampling phase, and switching to TypedVarInfo when each RV has been visited and all information about their types and distributions is known.

## 3 ADDITIONAL FUNCTIONALITIES FOR HANDLING PROBABILISTIC MODELS

In addition to the main functionalities of DynamicPPL, i.e., automatic type inference and tracing of model parameters, DynamicPPL provides various utility functions useful for handling probabilistic models and inference.

### 3.1 Contexts

To support tailored execution behaviour, DynamicPPL provides a variety of so-called contexts. Each model run happens in a specific context. The DefaultContext computes the logarithm of the joint probability of the observations and parameters. The LikelihoodContext is used to compute likelihoods. The PriorContext is used to compute the probability of some of the parameters given



the observations or ignoring the observations if possible. Lastly, the `MiniBatchContext` enables the scaling of the log likelihood term of the log joint probability to properly compute the stochastic gradient required in some variants of variational inference, such that it is in expectation equal to the full batch gradient.

### 3.2 Automatic Differentiation

`DynamicPPL`'s `VarInfo` type was written carefully to enable efficient interoperability with Julia's AD packages: `ForwardDiff.jl`[6] for vectorized forward-mode AD and `Tracker.jl`[7] for reverse-mode AD. We achieve this by specializing the types of the value vectors in `vi::TypedVarInfo` for the RVs which we would like to differentiate with respect to. This is then used in `Turing` to perform HMC sampling as well as HMC within Gibbs.

### 3.3 Early Rejection for Metropolis-style Algorithms

`DynamicPPL` allows the users to conditionally shortcut the model run in cases where the current set of parameters is known to have a zero probability. This can be used by the user to guard against numerical errors in the model. For example, calling functions like `log` or `sqrt` in the model with a negative argument will lead to a run-time error, hence the need for a mechanism for the early rejection of the samples. To reject a sample and quit the model run, users can overwrite the current log probability accumulator with `-Inf` (i.e. 0 probability) using `@logpdf() = -Inf` followed by a **return** statement to terminate the model execution.

### 3.4 Caching Expensive Computations

`DynamicPPL` currently does not provide a functionality to perform dependency analysis of variables in the model to avoid recomputing values that depend on constant variables during Gibbs sampling. This is a work in progress; we refer to [Gabler et al. 2019] for details. However, the Julia package Memoization.jl[8] can be used to create a memo for each expensive function in the model, leading to significantly faster Gibbs sampling of computationally expensive models[9]. Memoization can also help in cases where an intermediate variable in the model depends on RVs which can only take a small number of value combinations. Memoization will therefore create a dictionary mapping the inputs of the function to its outputs avoiding the re-computation of values that were computed before.

### 3.5 Probability Queries

Beside sampling, the infrastructure introduced above enables user-friendly probabilistic queries for a given model. For example, one can compute the likelihood of a new pair `(x, y)` given values for the linear model's regression coefficients `w` and the standard deviation `s` using:

`prob"X = [1.0, 2.0]', y = [2.0] | w = [0.5, 0.0], s = 1.0, model = linreg"`

This syntax is known as a string macro in Julia enabling us to parse an arbitrary DSL passed in the form of a string to generate Julia code at parse-time that the compiler then efficiently compiles. Other possible queries are:

`prob"w = [1.0, 1.0]', s = 1.0 | model = linreg"`
`prob"X = [1.0, 2.0]', y = [2.0], w = [0.0, 0.0], s = 1.0 | model = linreg"`

---

[6]https://github.com/JuliaDiff/ForwardDiff.jl
[7]https://github.com/FluxML/Tracker.jl
[8]https://github.com/marius311/Memoization.jl
[9]For an example, see e.g.
https://turing.ml/dev/docs/using-turing/performancetips#reuse-computations-in-gibbs-sampling



The first query computes the prior probability of the values given for w and s, while the second computes the joint probability for the given the model.

In addition, `DynamicPPL` also enables queries using an MCMC chain, an instance of the type `MCMCChain` from the Julia package `MCMCChains.jl`[10]. The following syntax computes the predictive posterior probability of an unseen data point.

```
prob"X = [1.0, 1.0]', y = [2.0] | chain = chain_instance"
```

The values for w and s are taken from the MCMC chain `chain_instance`, an instance of the type `MCMCChain`.

## 4 SOME BENCHMARKING RESULTS

To evaluate the computational performance of `DynamicPPL`, we run static HMC with 4 leapfrog steps for 2,000 iterations (step size varies for different models) on 8 standard benchmark models. All experiments are performed using `Turing` together with `AdvancedHMC` [Xu et al. 2019]. The run-time for all models is shown in Table 1.

|  | 10,000-D Gaussian | Gauss Unknown | Naive Bayes | Logistic Regression |
|---|---|---|---|---|
| Turing | **8.233 ± 0.144** | 2.152 ± 0.011 | **7.617 ± 0.063** | **4.225 ± 1.393** |
| Stan | 11.970 ± 0.445 | **0.349 ± 0.002** | 13.852 ± 0.054 | 60.932 ± 0.099 |
|  | Hierarchical Poisson | Sto. Volatility | Semi-sup HMM | LDA |
| Turing | 0.331 ± 0.011 | 62.159 ± 0.854 | 463.213 ± 26.045 | 378.762 ± 7.910 |
| Stan | **0.115 ± 0.047** | **0.705 ± 0.018** | **5.033 ± 0.058** | **43.888 ± 0.504** |

Table 1. Inference time (in seconds; smaller is better). Timings for Gaussian with unknown parameters (Gauss Unknown) is based on 10,000 observations with one dimension. Naive Bayes is based on 1,000 observations from MNIST projected onto 40 dimensions using principle component analysis. Logistic regression is based on 10,000 observations of 100 dimensions. Hierarchical Poisson is based on 50 observations. Stochastic volatility (Sto. Volatility) is based on 500 observations. Semi-supervised hidden Markov model (Semi-sup HMM) uses a five dimensional discrete latent space, a discrete 20 dimensional observation space and the timings are base on 300 observations where 200 are unsupervised. Latent Dirichlet allocation (LDA) is using a vocabulary size of 100, 5 topics and 10 documents with an average length of 1,000 words. See https://github.com/TuringLang/TuringExamples/tree/benchmarks for more details.

Overall, the performance of `DynamicPPL` compares favourably with `Stan`'s; `DynamicPPL` is more efficient on 3 out of 8 models while slower on the others. This is a surprisingly good result that we did not anticipate, since the `C++` implementation of `Stan` has been highly-optimised. For models with unfavourable run time compared to `Stan`, we have analysed the reasons. The findings suggest that the main reason is due to other libraries that `DynamicPPL` currently depends on and not `DynamicPPL` itself. In particular, for time series models (the stochastic volatility and the hidden Markov model), `Stan` is much faster than `DynamicPPL` because of the reverse-mode AD library used by `DynamicPPL`, `Tracker.jl`, which makes repeated use of Julia's dynamic dispatch leading to a large run-time overhead. In benchmarks where `DynamicPPL` came ahead of `Stan`, the overhead introduced by `Tracker.jl` was relatively small so these benchmarks more accurately represent the performance of `DynamicPPL`. Additionally, this AD issue can be worked around and improved in future versions of `Tracker.jl` and `DistributionsAD.jl`[11]; we are currently working on fixes.

---

[10]https://github.com/TuringLang/MCMCChains.jl
[11]https://github.com/TuringLang/DistributionsAD.jl



Supporting another reverse-mode AD Julia package like the promising Zygote.jl[12] [Innes et al. 2019] is also planned. We are optimistic that the performance of DynamicPPL can catch up with Stan in all considered models in the near future.

## 5 DISCUSSION AND FUTURE WORK

We presented the preliminary design and promising performance of DynamicPPL, a pure Julia library implementing a high-performance DSL for dynamic probabilistic models. DynamicPPL inherits and extends the DSL implementation in Turing.jl, and has became the default frontend for Turing.jl. Some notable properties of DynamicPPL include a focus on computational efficiency and modularity in the design, with the aim to support practical uses and probabilistic programming research. Another important advantage of DynamicPPL when compared to Stan is its full support for Julia's syntax, enabling direct access of excellent parallel computing, GPU-acceleration, and intensive numerical libraries available in Julia.

In the future, we hope to extend the current type inference capabilities of DynamicPPL for traces to support more modeling features such as: hierarchical/compositional modeling. Compositional modeling support would allow the use of a model in other models, enabling building complex models from simple parts like elementary distributions. Another future direction is to encode conditional independence relationships (see, e.g. [Bingham et al. 2019; Gabler et al. 2019; Mansinghka et al. 2014; Minka and Winn 2008; Murray and Schön 2018]) in some restricted modeling context which can be useful to implement BUGS-style [Lunn et al. 2000] Gibbs sampling, and Infer.net-style [Minka and Winn 2008] message passing algorithms.

---

[12]https://github.com/FluxML/Zygote.jl